\pdfoutput=1
\documentclass[11pt]{article}

\usepackage[margin=1in]{geometry}
\usepackage{authblk}
\usepackage[utf8]{inputenc}
\usepackage[T1]{fontenc}
\usepackage{lmodern}

\usepackage{amsmath,amsfonts,amssymb}
\usepackage{array}
\usepackage{booktabs}
\usepackage{graphicx}
\usepackage{tikz}
\usepackage{pgfplots}
\usepackage{textcomp}
\usepackage{xurl}
\makeatletter
\@ifundefined{if@mainmatter}{\newif\if@mainmatter\@mainmattertrue}{}
\makeatother
\usepackage[natbibapa]{apacite}
\usepackage{hyperref}
\renewcommand{\doiprefix}{}
\def\refdoi#1{\url{#1}}
\AtBeginDocument{}
\setcitestyle{authoryear,round,aysep={,},yysep={,}}
\newcommand{\keywords}[1]{\par\medskip\noindent\textbf{Keywords:} {\def\and{\unskip\,$\cdot$\,}#1}}

\hypersetup{
  hidelinks,
  pdfauthor={Quy P. Nguyen, Duong Q. Nguyen, Dinh P. T. Nguyen, Truong-Giang T. Le, Truong A. Bui, Vi H. Vo, Quang-Thinh Bui, My-Phuong Ngo, Trong N. Do, Thao P. Le, Phuc M. Nguyen, H. Nguyen-Xuan},
  pdftitle={VeriMath: Human-in-the-Loop Generation and Verification of Curriculum-Aligned Mathematics Assessment Items}
}

\usetikzlibrary{patterns}
\pgfplotsset{compat=1.18}

\newcounter{algcounter}

\newenvironment{verimathalgorithm}[1]{%
  \refstepcounter{algcounter}%
  \begin{figure}[!t]%
  \centering%
  \begin{minipage}{0.96\linewidth}%
  \hrule
  \vspace{2pt}%
  \noindent\textbf{Algorithm \thealgcounter. #1}\par%
  \vspace{2pt}%
  \hrule
  \vspace{4pt}%
  \small%
}{%
  \vspace{4pt}%
  \hrule%
  \end{minipage}%
  \end{figure}%
}

\begin{document}
\shortcites{anderson2001taxonomy,cheng2025eqpr,christ2024mathwell,christ2026edumath,cobbe2021training,haladyna2002review,kibar2026instructional,karbasi2025multiagent,kasneci2023chatgpt,kurdi2020systematic,lightman2024lets,liu2023g,lu2024mathgenie,luo2023wizardmath,manem2025sandmath,cassano2026instructors,niyarepola2022math,pei2025mathfusion,polozov2015personalized,qin2023mathematical,qin2024math,shaden2026mathnet,toshniwal2024openmath2,utami2024contextualized,verschaffel2020word,wang2021math,wang2023selfconsistency,wei2022chain,yu2024metamath,zheng2023judging,zhou2019towards}

\title{VeriMath: Human-in-the-Loop Generation and Verification of Curriculum-Aligned Mathematics Assessment Items}
\author[1]{Quy P. Nguyen\thanks{First author. Email: \href{mailto:quy.np@vinuni.edu.vn}{quy.np@vinuni.edu.vn}}}
\author[1]{Duong Q. Nguyen}
\author[2]{Dinh P. T. Nguyen}
\author[3]{Truong-Giang T. Le}
\author[3]{Truong A. Bui}
\author[4]{Vi H. Vo}
\author[5]{Quang-Thinh Bui}
\author[6]{My-Phuong Ngo}
\author[1]{Trong N. Do}
\author[7]{Thao P. Le}
\author[8]{Phuc M. Nguyen}
\author[1,9]{H. Nguyen-Xuan\thanks{Corresponding author. Email: \href{mailto:hung.nx2@vinuni.edu.vn}{hung.nx2@vinuni.edu.vn}}}
\affil[1]{Center for AI Research, VinUniversity, Ho Chi Minh City, Viet Nam}
\affil[2]{An Lao High School, Gia Lai, Viet Nam}
\affil[3]{Chu Van An High School for Gifted Students, Gia Lai, Viet Nam}
\affil[4]{FPT Play, Ho Chi Minh City, Viet Nam}
\affil[5]{Applied Mathematics and Data Science (AMaDaS) Research Group, Tien Giang University, Dong Thap, Viet Nam}
\affil[6]{Tien Giang High School for the Gifted, Dong Thap, Viet Nam}
\affil[7]{Department of Mathematics and Statistics, Quy Nhon University, Quy Nhon City, Viet Nam}
\affil[8]{College of Electro-mechanics, Construction and Agro-forestry of Central Vietnam, Gia Lai, Viet Nam}
\affil[9]{College of Engineering and Computer Science, VinUniversity, Hanoi, Viet Nam}
\date{}
\maketitle

\begin{abstract}

Automatic generation of mathematics assessment items is a growing area in educational artificial intelligence, yet classroom deployment remains challenging. Generated questions must satisfy multiple constraints: mathematical correctness, curriculum alignment, appropriate cognitive difficulty, and educator reviewability. While prior work has advanced text generation, reliable assessment authoring requires granular control over localized examination formats and expert educational judgment. To address this challenge, we present VeriMath, a generalized, human-in-the-loop framework for generating rigorous, curriculum-constrained mathematics assessment drafts. Using annotated reference items, the framework anchors generation to the requested examination section and cognitive reasoning level, then applies sequential automated verification and guided repair before submitting drafts for expert review. We evaluate VeriMath through a national curriculum case study using 4,555 mathematics items from the Vietnamese Grade 12 curriculum and 50 distinct generation requirements covering multiple-choice, true/false, and short-answer formats. Across two model backbones, the framework produces verifier-accepted drafts for 94\% (47/50) of requirements per run. Blind expert evaluations confirm high mathematical accuracy and curriculum alignment, with strict teacher-readiness assessments accepting up to 56.8\% of generated items. Analysis of rejected drafts reveals persistent domain-agnostic challenges involving formatting, statement validity, and solution-step completeness. These findings demonstrate that VeriMath can bridge generative AI and dependable educational applications by producing highly reviewable assessment drafts within tightly constrained national curriculum frameworks.

\keywords{Automatic question generation \and curriculum-aligned assessment \and educational artificial intelligence \and human-in-the-loop systems \and large language models \and mathematics assessment}

\end{abstract}

\section{Introduction}

Mathematics assessment items are central to classroom practice, formative feedback and examination preparation. They connect curriculum objectives, cognitive demand, response format, scoring evidence and learner interpretation. Educational taxonomies and item-writing guidelines show that assessment quality depends on alignment among learning goals, task demands and scoring decisions, not only on grammatical clarity \citep{anderson2001taxonomy,haladyna2002review}. In mathematics education, this alignment is especially demanding because a problem statement, its representations, its solution path and its answer format must remain mathematically coherent for the intended learners \citep{verschaffel2020word}.

Automatic question generation has long been studied as a way to reduce the cost of creating learning and assessment material. Systematic reviews describe a broad literature that includes template-based, rule-based, semantic and neural methods for educational question generation \citep{kurdi2020systematic}. Recent instructional design research identifies pedagogical alignment and content reliability as persistent concerns in authoring supported by artificial intelligence (AI) \citep{kibar2026instructional}. In mathematics, early and intermediate systems generated personalized or constrained word problems from structured specifications, equations, topics and context constraints \citep{polozov2015personalized,zhou2019towards,wang2021math}. More recent retrieval-based and memory-based approaches improve relevance and diversity by reusing related examples while preserving generation flexibility \citep{qin2024math}.

Large language models (LLMs) extend this line of work because they can produce fluent explanations, adapt content to learner contexts and support educational content creation at scale. Research on large language models in education emphasizes opportunities for personalization and teacher support, but also stresses the need for critical checking, pedagogical grounding and human oversight \citep{kasneci2023chatgpt}. At the assessment-authoring level, \citet{cassano2026instructors} embed item-writing criteria in prompts for transcript-based multiple-choice question generation and evaluate outputs through instructors' judgments. These findings establish evidence-informed validation and human judgment as necessary quality controls for educational use.

Mathematics-specific studies further use generative models to create contextualized and personalized word problems grounded in authentic contexts \citep{utami2024contextualized}. Teacher-evaluated systems such as MATHWELL and EDUMATH show that solvability and answer accuracy do not by themselves ensure educational appropriateness or standards alignment \citep{christ2024mathwell,christ2026edumath}. These developments motivate moving from isolated prompt outputs toward authoring workflows in which generation, verification and expert judgment are explicitly connected.

The problem studied here is curriculum-constrained mathematics assessment item authoring. Given a target grade, curriculum learning outcome, assessment schema, response constraints and cognitive or difficulty specification, the goal is to generate an original item draft with a correct solution and usable metadata. The draft must be mathematically valid, aligned with the intended learning outcome, matched to the requested level, structurally compatible with the target item format, \LaTeX{}-compilable and sufficiently novel with respect to prior assessment material. This formulation is broader than ordinary math word problem generation because it treats an item as an assessment artifact that must be ready for professional review.

Existing methods leave several limitations for this setting. The instructional-design review emphasizes practitioner judgment, while the transcript-based question-generation study illustrates quality-constrained authoring and instructor evaluation \citep{kibar2026instructional,cassano2026instructors}. For curriculum-constrained mathematics assessment, the corresponding design problem is to connect local outcomes, assessment schemas, difficulty conventions, solution verification, novelty and teacher readiness within one workflow. Template and symbolic systems can enforce structure, but they often provide limited linguistic and mathematical diversity. Neural and large language model-based systems improve fluency and variation, but they may still produce hidden assumptions, invalid official answers, ambiguous wording, curriculum mismatches or near copies of familiar items \citep{wang2021math,christ2024mathwell,utami2024contextualized}. Synthetic mathematical data pipelines such as MathGenie, MetaMath, WizardMath and OpenMathInstruct 2 demonstrate the value of generated mathematics data for training reasoning models, but their primary evaluation target is often downstream solver performance rather than teacher readiness for classroom or examination use \citep{lu2024mathgenie,yu2024metamath,luo2023wizardmath,toshniwal2024openmath2}.

These limitations motivate three research questions: (RQ1) Can large language models generate structurally complete mathematics assessment drafts under explicit curriculum, format and level constraints? (RQ2) Can a context-specific teacher-annotated corpus provide useful anchors for difficulty and style without encouraging surface reuse? (RQ3) How large is the remaining gap between automated verification and teacher judgments of readiness? These questions are important because educational deployment requires more than fluent generation. It requires a workflow that makes generated items controllable, inspectable and open to expert rejection.

VeriMath addresses these questions through a configurable human-in-the-loop authoring framework for mathematics assessment drafts. A deployment supplies its assessment schemas, curriculum outcomes, cognitive or difficulty conventions, response constraints, annotated reference corpus and expert-review rubric. The framework retrieves reference items that match the requested schema and level specification. These references guide style, format and reasoning depth, while the generated item must remain original. A generator then produces a complete draft, and repeated verification checks mathematical correctness, requirement alignment, answer consistency and structural validity. When verification fails, the failure report is used to revise the draft under the same requirement. The final output is not treated as an official examination item; it is a candidate for teacher review.

The framework is therefore defined independently of any particular national curriculum, whereas each deployment is configured with context-specific educational resources and conventions. We instantiate and evaluate it through a Vietnamese Grade 12 national-curriculum case study. The reference corpus contains 4,555 teacher-annotated national and practice examination questions, and the experimental requirement set covers multiple-choice, true-or-false and short-answer formats. Two large language model backbones are evaluated through full-pipeline diagnostics and blind expert review. The central empirical issue is the difference between automated acceptance and teacher readiness, because this difference indicates which defects still require expert assessment judgment after automated verification. The study provides evidence from this single curriculum context rather than a cross-country validation.

The main contributions are as follows:
\begin{itemize}
\item A curriculum-constrained formulation of mathematics assessment item authoring in which assessment schemas, learning outcomes, cognitive or difficulty specifications, response constraints and curriculum metadata are supplied by the deployment context.
\item A configurable human-in-the-loop framework that combines context-specific difficulty anchoring with repeated verifier-guided revision before a draft is accepted for expert review.
\item A national-curriculum case-study instantiation using a 4,555-item Vietnamese Grade 12 mathematics corpus annotated by teachers for assessment schema and cognitive or difficulty level.
\item An empirical evaluation of this instantiation with two large language model backbones, including automated acceptance, runtime and token cost, embedding-based novelty, blind expert review and expert-marked error categories.
\item An analysis of the readiness gap between automated verification and teacher judgment, identifying remaining failure modes and the continuing need for human oversight in constrained educational generation.
\end{itemize}

\section{Related Work}

Research on automated mathematics item creation has developed along three connected directions. The first direction studies automatic question generation and mathematical word problem generation. It moves from structured generation and knowledge sources toward neural generation with stronger mathematical constraints. The second direction uses large language models for controllable educational content generation, where teacher annotations, standards and authentic contexts become central quality signals. The third direction studies verification and evaluation, including reasoning verifiers, synthetic mathematical data pipelines and model as judge rubrics. VeriMath draws from all three directions, but it targets a stricter authoring problem in which local curriculum metadata, item format, difficulty anchoring, novelty and teacher readiness must be considered together. Table \ref{tab:method_positioning} summarizes how VeriMath is positioned relative to major methodological lines in mathematics item generation.

\begin{table}[!t]
\caption{Positioning of VeriMath Relative to Major Lines of Mathematics Item Generation}
\label{tab:method_positioning}
\centering
\scriptsize
\setlength{\tabcolsep}{3.2pt}
\renewcommand{\arraystretch}{1.18}
\begin{tabular}{p{0.15\textwidth}p{0.22\textwidth}p{0.22\textwidth}p{0.30\textwidth}}
\toprule
\textbf{Research line} & \textbf{Representative work} & \textbf{Main technical idea} & \textbf{Strengths and remaining gap} \\
\midrule
Structured generation & \citet{polozov2015personalized} & Generate problems from structured specifications, mathematical models or student relevant contexts. & Strong control over formal structure, but limited variation and weak support for local examination formats. \\
Neural word problem generation & \citet{zhou2019towards}, \citet{wang2021math}, \citet{niyarepola2022math} & Condition generation on equations, topics, contexts or language resources. & Improves fluency and diversity, but does not fully address curriculum metadata, item schema validity or expert readiness. \\
Pedagogically guided educational authoring & \citet{kibar2026instructional,cassano2026instructors} & Connect AI-assisted design with pedagogical review and explicit item-writing criteria. & Provides an instructional-design and assessment-authoring basis; VeriMath examines these concerns in curriculum-constrained mathematics. \\
Planning and retrieval & \citet{qin2023mathematical,qin2024math}, \citet{cheng2025eqpr} & Use structure planning, memory retrieval or question planning to guide generation. & Improves controllability, yet generated outputs still require stronger review for mathematical validity and assessment use. \\
Teacher evaluated generation & \citet{christ2024mathwell,christ2026edumath}, \citet{utami2024contextualized} & Use teacher labels, standards or authentic contexts as educational quality signals. & Connects generation to pedagogy, but many systems focus on word problems rather than national examination item schemas. \\
Verification and evaluation & \citet{cobbe2021training}, \citet{lightman2024lets}, \citet{liu2023g}, \citet{zheng2023judging} & Use verifier models, process feedback or model based rubrics to screen generated reasoning. & Scales diagnostic review, but automatic judgment remains an imperfect substitute for expert educational assessment. \\
\textbf{VeriMath} & This study & Combine context-specific difficulty anchors, repeated verifier guided revision and blind teacher review. & Targets curriculum constrained assessment drafts with configurable schemas and explicit level, novelty and readiness evidence. \\
\bottomrule
\end{tabular}
\end{table}

\subsection{Automatic Question Generation and Mathematical Word Problem Generation}
Automatic question generation has long been studied as an educational technology for producing learning objects, practice questions and assessment material. \citet{kurdi2020systematic} provide a broad foundation for educational question generation and show that evaluation practices vary across domains and generation settings. Within instructional design, \citet{kibar2026instructional} synthesize 28 studies and identify content reliability, pedagogical alignment and practitioner judgment as central considerations. In mathematics education, the generated item must also preserve conceptual coherence, valid representations and solvable structure \citep{verschaffel2020word}. General assessment research further shows that item quality depends on alignment among objectives, cognitive demand, response format and scoring interpretation \citep{anderson2001taxonomy,haladyna2002review}. These foundations motivate treating mathematical item generation as an assessment authoring problem rather than a surface text generation problem.

Early mathematical word problem generation emphasized explicit control. \citet{polozov2015personalized} generate personalized word problems from structured specifications, mathematical models and student relevant contexts. This type of method is attractive because it exposes variables and constraints to the author. Its main limitation is that hand designed structures restrict linguistic variety and make adaptation to complex examination formats costly.

Neural generation increases diversity but introduces new validity challenges. \citet{zhou2019towards} generate mathematical word problems from equations and topics using a neural model with a fusion mechanism and an entity consistency objective. \citet{wang2021math} improve problem generation with mathematical consistency and context constraints. \citet{niyarepola2022math} extend the task to multilingual generation, showing that language resources matter for broader educational access. Qin \textit{et al.} introduce structure planning and knowledge enhancement to improve logical coherence \citeyearpar{qin2023mathematical}, and later use disentangled memory retrieval to retrieve related logical and scenario descriptions for generation \citeyearpar{qin2024math}. These works establish important methods for controlling mathematical content, but they mainly focus on word problem quality rather than complete assessment artifacts.

Recent work also connects question generation with instructors' evaluation of usable assessment material. \citet{cassano2026instructors} build a transcript-based multiple-choice question workflow around item-writing guidelines and report a mixed-methods evaluation with seven instructors and 257 questions. Remaining relevance and wording issues illustrate why structural compliance alone is insufficient. Their study concerns video-based materials rather than national-curriculum mathematics. VeriMath focuses on the latter setting through teacher-annotated anchors, explicit schemas, iterative verification and repair, novelty screening and blind expert review.

\subsection{Controllable and Teacher Evaluated Educational Generation}
Large language models broaden the design space because they can generate fluent explanations, adapt wording to educational contexts and support authoring at scale. At the same time, education focused studies emphasize that generative artificial intelligence requires human oversight, pedagogical design and careful integration with teacher judgment \citep{kasneci2023chatgpt}. This perspective is important for mathematics assessment because a fluent generated question may still be misaligned with the intended learning outcome or unfair for the target learners.

Recent work therefore moves from generic generation toward controllable and teacher evaluated generation. \citet{utami2024contextualized} study contextualized and personalized mathematical word problem generation in authentic contexts and examine its influence on geometry learning. MATHWELL uses teacher annotations to target solvability, answer accuracy and educational appropriateness for mathematics word problems \citep{christ2024mathwell}. EDUMATH extends this direction toward standards aligned and customized educational word problems with expert and model based evaluation \citep{christ2026edumath}. These systems show that educational quality must be evaluated with pedagogical criteria, not only answer correctness.

Another line improves controllability through planning, collaboration and difficulty manipulation. \citet{cheng2025eqpr} formulate educational mathematical question generation as a planning problem with reflection over objectives. \citet{karbasi2025multiagent} use a collaborative multi agent framework to refine generated math problems under criteria such as clarity, relevance, answerability and difficulty matching. SAND Math generates novel mathematics problems and modifies their difficulty for synthetic training data construction \citep{manem2025sandmath}. These approaches are methodologically close to VeriMath because they introduce intermediate control signals. VeriMath differs by using context-specific teacher annotated difficulty anchors, enforcing configurable assessment schemas and measuring readiness through blind expert review.

\subsection{Verification, Synthetic Data and Evaluation}
Verification is a central issue because mathematical outputs can look plausible while containing invalid assumptions, wrong answers or incomplete solutions. Chain of thought prompting encourages intermediate reasoning, and self consistency aggregates multiple reasoning paths to improve answer reliability \citep{wei2022chain,wang2023selfconsistency}. \citet{cobbe2021training} show that trained verifiers can rank candidate solutions for grade school mathematics problems. \citet{lightman2024lets} show that process supervision can provide feedback at intermediate reasoning steps and improve mathematical reasoning reliability. These studies focus mainly on solving or judging solutions. VeriMath adapts the verification idea to item authoring, where the object under review includes the problem statement, answer, solution, metadata and format.

Generated mathematics is also used to train stronger reasoning models. MetaMath bootstraps mathematical questions for large language model training \citep{yu2024metamath}. WizardMath uses reinforced instruction evolution for mathematical reasoning \citep{luo2023wizardmath}. MathGenie uses question back translation and rationale based verification when constructing synthetic data \citep{lu2024mathgenie}. OpenMathInstruct 2 scales open mathematical instruction data, while MathFusion combines mathematical instructions across sources to improve problem solving \citep{toshniwal2024openmath2,pei2025mathfusion}. These methods demonstrate the value of generated mathematics data, but their primary success criterion is downstream solver performance. Assessment item authoring requires a different criterion, namely whether a teacher can safely review and deploy the generated item.

Finally, model based evaluation offers scalable diagnostics but remains limited for educational readiness. G Eval uses large language models with generated evaluation steps and form based scoring for natural language generation quality \citep{liu2023g}. \citet{zheng2023judging} study model as judge evaluation and document useful agreement with human preferences as well as biases such as position, verbosity and self enhancement effects. These findings motivate using automated diagnostics cautiously. VeriMath therefore reports verifier acceptance and operational metrics, but treats blind expert review as the primary evidence for educational readiness.

\section{Task and Dataset}
\label{sec:task-dataset}

\subsection{Task Formulation and Output Schema}
\label{sec:task-formulation}
Let $R=(s,o,L,q,m)$ denote an item generation requirement. The variable $s\in\mathcal{S}$ identifies an assessment schema from a set $\mathcal{S}$ defined by the deployment context, $o$ is the curriculum learning outcome or standard, $L$ is the context-specific cognitive or difficulty specification, $q$ contains response constraints such as answer type and language and $m$ contains curriculum metadata such as grade, subject, chapter and lesson section. Neither $\mathcal{S}$ nor the taxonomy underlying $L$ is fixed by the VeriMath architecture. A deployment supplies its schema definitions and operational level conventions as part of the framework configuration.

Given $R$ and a reference corpus $\mathcal{D}_{\mathrm{ref}}$, the goal is to generate an assessment draft $X=(P,S,M)$, where $P$ is the problem statement, $S$ is the solution and official answer and $M$ is the generated metadata. The tuple $X$ should be mathematically correct, aligned with $o$, matched to $L$, valid under the requested assessment schema and sufficiently novel with respect to $\mathcal{D}_{\mathrm{ref}}$. The output is treated as a teacher reviewable draft rather than a final examination artifact.

In the Vietnamese Grade 12 case-study instantiation, $\mathcal{S}_{\mathrm{VN}}=\{1,2,3\}$ uses three schema identifiers mapped to the item formats summarized in Table \ref{tab:exam_section_schema}. The multiple-choice and short-answer schemas use a single level, while the true-or-false cluster schema uses $L=(\ell_1,\ell_2,\ell_3,\ell_4)$ because it contains four independently judged statements. The response structure for this instantiation is determined by $s$ and enforced by the generation and verification process described in Section \ref{sec:methodology}. These three schemas are the configuration evaluated in this study, not a universal schema set required by VeriMath.

\begin{table}[!t]
\caption{Assessment Schemas Used in the Vietnamese Case-Study Instantiation}
\label{tab:exam_section_schema}
\centering
\small
\renewcommand{\arraystretch}{1.15}
\begin{tabular}{p{0.10\textwidth} p{0.16\textwidth} p{0.38\textwidth} p{0.20\textwidth}}
\toprule
\textbf{Case-study code $s$} & \textbf{Item format} & \textbf{Required output} & \textbf{Level specification} \\
\midrule
1 & Multiple choice & One stem, four answer options, one correct label and a detailed solution. & One level from Table \ref{tab:cognitive_levels}. \\
2 & True or false cluster & One shared stem, four statements, four binary labels and a detailed solution for all statements. & Four entry profile $L=(\ell_1,\ell_2,\ell_3,\ell_4)$. \\
3 & Short answer & One stem, one numeric final answer and a detailed solution, with rounding instructions when needed. & One level from Table \ref{tab:cognitive_levels}. \\
\bottomrule
\end{tabular}
\end{table}

\subsection{Reference Corpus, Annotation and Difficulty Levels}
\label{sec:reference-corpus}
The case-study reference corpus $\mathcal{D}_{\mathrm{ref}}$ contains 4,555 mathematics problems collected from Vietnamese national and practice examination sources. Mathematics teachers annotated each item with its assessment schema and cognitive or difficulty level. The corpus is used to select difficulty anchors, estimate novelty by maximum embedding similarity and define the expected distribution of topics and formats. To reduce leakage risk, anchors are used only as format and difficulty references, while generated items are checked for novelty against the full corpus.

The annotation protocol is designed as a two stage process: initial labeling followed by senior teacher adjudication of disagreements. Discussion focuses on the minimum reasoning depth needed to solve the item rather than on surface complexity. The resulting labels are teacher operational categories for anchor selection and level matching. They are not treated as psychometrically calibrated item difficulty parameters.

VeriMath does not require a universal difficulty taxonomy; it assumes that each deployment supplies an operational specification for $L$ and compatible corpus annotations. In the Vietnamese case study, we use one variable for both cognitive level and practice-examination difficulty. This reflects the local authoring setting, where perceived difficulty depends on reasoning depth, item format and examination convention. The three levels used in this instantiation are defined in Table \ref{tab:cognitive_levels}.

\begin{table}[!t]
\caption{Teacher-Annotated Cognitive/Difficulty Levels in the Vietnamese Case Study}
\label{tab:cognitive_levels}
\centering
\small
\renewcommand{\arraystretch}{1.15}
\begin{tabular}{p{0.27\linewidth}p{0.61\linewidth}}
\toprule
\textbf{Level} & \textbf{Definition} \\
\midrule
Recognition & Direct formula or concept recall with at most one reasoning step. \\
Comprehension & Requires interpretation and combination of known concepts. \\
Application & Requires multi step reasoning or a non routine strategy. \\
\bottomrule
\end{tabular}
\end{table}

For expert evaluation, generated items are reviewed under a blind protocol. Mathematics educators rate each item using four criteria: mathematical correctness and answer consistency, alignment with the assigned learning outcome, match to the required cognitive or difficulty level and clarity or readiness for examination use. Correctness and alignment are recorded as pass or fail labels. Level match and clarity are scored on a 1 to 5 scale. Generation source, model identity and process traces are hidden from reviewers.

\subsection{Requirement Set and Evaluation Dimensions}
\label{sec:requirement-set}
The experimental requirements are constructed from the Vietnamese Grade 12 mathematics curriculum, with emphasis on learning outcomes relevant to the national high school graduation examination. Each requirement specifies one of the case-study assessment schemas, a curriculum learning outcome, a cognitive or difficulty level and topic metadata. The study is restricted to Grade 12 because this grade is most directly aligned with the examination setting considered here.

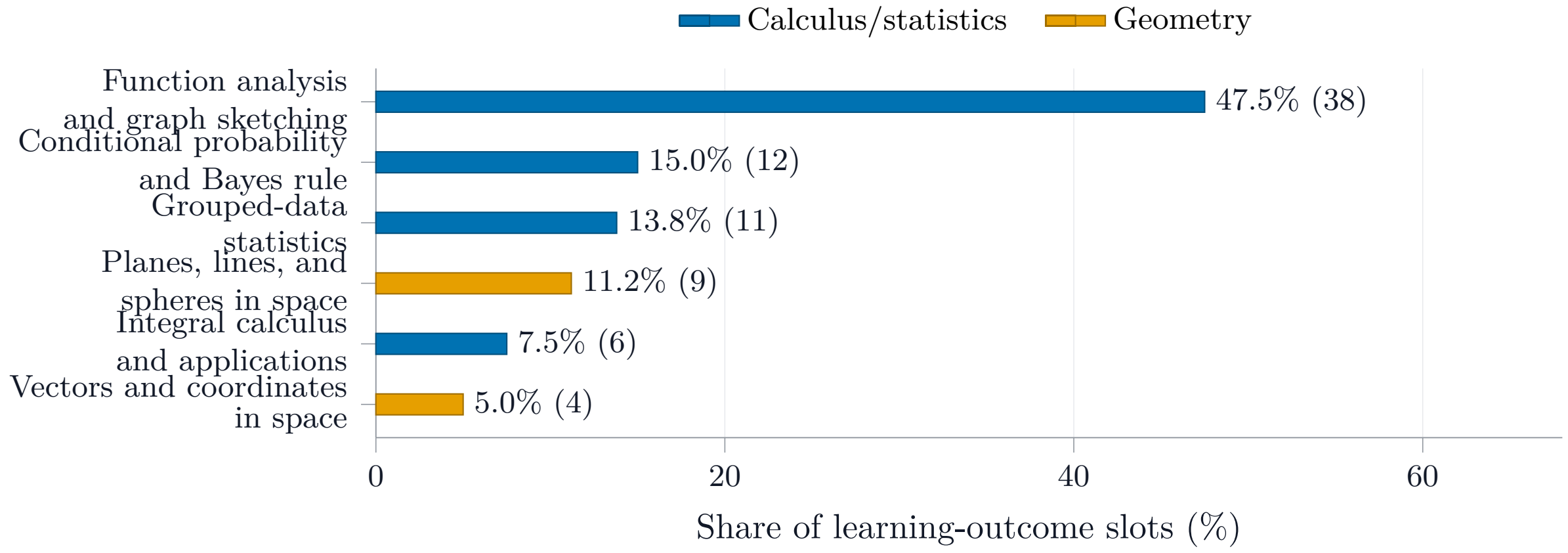
\begin{figure}[!t]
\centering

\definecolor{vmTopicBlue}{HTML}{0072B2}
\definecolor{vmTopicAmber}{HTML}{E69F00}
\definecolor{vmTopicBorder}{HTML}{374151}
\definecolor{vmTopicGrid}{HTML}{E5E7EB}
\definecolor{vmTopicText}{HTML}{111827}

\pgfplotsset{
  vm topic axis/.style={
    axis x line*=bottom,
    axis y line*=left,
    axis line style={vmTopicBorder!55, line width=0.35pt},
    tick style={vmTopicBorder!55, line width=0.3pt},
    tick align=outside,
    xmajorgrids=true,
    grid style={solid, vmTopicGrid, line width=0.25pt},
    tick label style={font=\footnotesize, text=vmTopicText},
    label style={font=\small, text=vmTopicText},
    y tick label style={font=\footnotesize, text=vmTopicText, align=right},
    every node near coord/.append style={
      font=\footnotesize,
      text=vmTopicText,
      anchor=west,
      xshift=2pt,
      inner sep=1pt
    },
    clip=false,
  },
}

\begin{tikzpicture}
\begin{axis}[
  vm topic axis,
  width=0.82\columnwidth,
  height=5.3cm,
  xmin=0,
  xmax=68,
  xtick={0,20,40,60},
  xlabel={Share of learning-outcome slots (\%)},
  ymin=0.45,
  ymax=6.55,
  ytick={1,2,3,4,5,6},
  yticklabels={
    {\shortstack[r]{Vectors and coordinates\\in space}},
    {\shortstack[r]{Integral calculus\\and applications}},
    {\shortstack[r]{Planes, lines, and\\spheres in space}},
    {\shortstack[r]{Grouped-data\\statistics}},
    {\shortstack[r]{Conditional probability\\and Bayes rule}},
    {\shortstack[r]{Function analysis\\and graph sketching}}
  },
  legend columns=2,
  legend style={
    draw=none,
    fill=none,
    font=\footnotesize,
    at={(0.5,1.05)},
    anchor=south,
    /tikz/every even column/.append style={column sep=0.8em}
  },
]
\addplot+[
  xbar,
  mark=none,
  bar width=6pt,
  fill=vmTopicBlue,
  draw=vmTopicBlue!70!black,
  nodes near coords={\pgfplotspointmeta},
  point meta=explicit symbolic,
]
coordinates {
  (47.5,6) [47.5\% (38)]
  (15.0,5) [15.0\% (12)]
  (13.8,4) [13.8\% (11)]
  (7.5,2) [7.5\% (6)]
};
\addlegendentry{Calculus/statistics}

\addplot+[
  xbar,
  mark=none,
  bar width=6pt,
  fill=vmTopicAmber,
  draw=vmTopicAmber!70!black,
  nodes near coords={\pgfplotspointmeta},
  point meta=explicit symbolic,
]
coordinates {
  (11.2,3) [11.2\% (9)]
  (5.0,1) [5.0\% (4)]
};
\addlegendentry{Geometry}
\end{axis}
\end{tikzpicture}

\caption{Topic-block distribution of the Grade 12 requirement set. Bars report the share of the 80 learning-outcome slots, with slot counts shown in parentheses}
\label{fig:input_topic_distribution}
\end{figure}

The requirement set contains 50 items sampled with a fixed schema ratio of $6{:}2{:}2$: 30 multiple choice items, 10 true or false clusters and 10 short answer items. The single outcome schemas in Table \ref{tab:exam_section_schema} each represent one curriculum learning outcome. The true or false cluster schema in Table \ref{tab:exam_section_schema} represents four learning outcome positions because every cluster contains four independent statements. Therefore, the 50 items collectively represent 80 curriculum learning outcome slots. The set contains 32 Recognition, 24 Comprehension and 24 Application slots. Table \ref{tab:input_subject_distribution} summarizes the subject distribution. Fig. \ref{fig:input_topic_distribution} shows the topic block distribution over these slots.

\begin{table}[!t]
\caption{Subject Distribution of the Grade 12 Requirement Set}
\label{tab:input_subject_distribution}
\centering
\small
\renewcommand{\arraystretch}{1.12}
\begin{tabular}{lcc}
\toprule
\textbf{Curriculum subject} & \textbf{Slots} & \textbf{Share} \\
\midrule
Calculus/statistics & 67 & 83.8\% \\
Geometry & 13 & 16.2\% \\
\bottomrule
\end{tabular}
\end{table}

Table \ref{tab:dataset_stats} summarizes the corpus and experimental pools.

\begin{table}[!t]
\caption{Dataset Statistics}
\label{tab:dataset_stats}
\centering
\small
\renewcommand{\arraystretch}{1.15}
\begin{tabular}{p{0.42\linewidth}cccc}
\toprule
\textbf{Data split or pool} & \textbf{Total} & \textbf{$s=1$} & \textbf{$s=2$} & \textbf{$s=3$} \\
\midrule
Reference corpus $\mathcal{D}_{\mathrm{ref}}$ & 4,555 & 2,988 & 449 & 1,118 \\
Experiment requirements & 50 & 30 & 10 & 10 \\
Experiment learning outcome slots & 80 & 30 & 40 & 10 \\
\bottomrule
\end{tabular}
\end{table}

Evaluation separates operational validity from teacher readiness. Operational diagnostics measure whether the pipeline can produce a structurally complete and verifier accepted item draft under $R$. Novelty measures dissimilarity from $\mathcal{D}_{\mathrm{ref}}$ using maximum embedding similarity. Teacher readiness is assessed separately through blind expert review. This separation prevents automatic acceptance or novelty scores from being interpreted as substitutes for educational quality.

\section{Methodology}
\label{sec:methodology}

\begin{figure}[!t]
\centering
\includegraphics[width=0.92\textwidth]{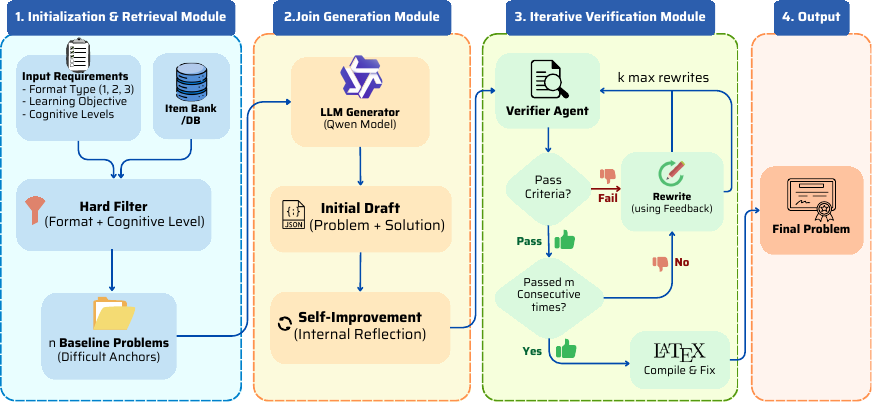}
\caption{Overview of the VeriMath framework. Difficulty Anchoring selects schema- and level-matched reference items, while CVM repeatedly verifies and repairs a candidate before acceptance}
\label{fig:framework}
\end{figure}

Using the task notation from Section \ref{sec:task-formulation}, VeriMath generates teacher reviewable mathematics assessment drafts under curriculum, schema and difficulty constraints supplied by the deployment context. It contains two main inference time components: Difficulty Anchoring, described in Section \ref{sec:difficulty-anchoring}, and a Consecutive Verification Mechanism (CVM), described in Section \ref{sec:cvm}. Fig. \ref{fig:framework} summarizes the framework.

\subsection{Difficulty Anchoring and Candidate Generation}
\label{sec:difficulty-anchoring}
Given a requirement $R=(s,o,L,q,m)$ with $s\in\mathcal{S}$, VeriMath selects $k$ anchor problems
\[
B_R=\{b_1,\ldots,b_k\}
\]
from $\mathcal{D}_{\mathrm{ref}}$. Anchor selection begins by filtering problems with the same assessment schema $s$ and a compatible level specification $L$, according to the rules supplied for the deployment. In the Vietnamese case-study instantiation, compatibility for the single-level schemas in Table \ref{tab:exam_section_schema} means the same level. For the true-or-false cluster schema, the selector uses three deterministic tiers over the four-entry profile $L$: exact profile matching, same level-distribution matching and required-level-presence matching. The last tier requires an anchor to contain each level that appears in the target profile at least once. If no tier yields enough anchors, the requirement is rejected rather than using a weaker schema-only search.

The selected anchors are not copied into the generated question. They provide examples of expected format, reasoning depth and examination style. The generator $\mathcal{M}_{\mathrm{gen}}$ receives $R$ and $B_R$ and produces an initial candidate
\[
X_0=(P_0,S_0,M_0)
\]
following the output schema associated with $s$. A self improvement step then checks the candidate for mathematical mistakes, formatting issues and mismatches with $L$. The resulting candidate $X_{\mathrm{cand}}=(P_{\mathrm{cand}},S_{\mathrm{cand}},M_{\mathrm{cand}})$ is sent to CVM.

The generator, verifier and corrector use different role prompts. The generator acts as an examination item writer. The verifier acts as a strict mathematics teacher and reviewer. The corrector revises an item only after receiving a detailed failure trace. For the true or false cluster schema in Table \ref{tab:exam_section_schema}, both generator and verifier receive the four entry level profile $L$, and the verifier is instructed to reject an item if any statement fails to match its required level.

\noindent\textbf{Representative generator prompt.}
\begin{quote}\small
You are an expert mathematics assessment item writer. Generate one original mathematics problem that satisfies the required educational competencies and follows the specified structural format. The problem must test the required competency, be objectively gradable, be mathematically correct, match the reasoning depth of the provided anchor problems and avoid paraphrasing those anchors. Use valid \LaTeX{} syntax, avoid ambiguous wording and return one valid JSON object following the specified assessment schema.
\end{quote}

The corresponding user message specifies the curriculum learning outcome, the selected difficulty anchor problems and the instruction to generate a new original problem that matches the reasoning depth of the anchors while remaining precise and unambiguous.

\subsubsection{Requirement Output Examples With Anchors}
\label{sec:anchor-examples}
The following examples come from the Vietnamese case-study instantiation and show the observable requirement output behavior of the generator without intermediate verifier and correction steps. In each case, the requirement specifies an assessment schema, target level and learning outcome. Anchors are positive references for format and reasoning depth, but negative references for content reuse.

\begin{quote}\small
\textbf{Multiple choice example.} The requirement uses the $s=1$ schema in Table \ref{tab:exam_section_schema}. It asks for one multiple choice item with four answer options, one correct label and a detailed solution. The target level is Recognition and the knowledge point is the basic substitution method. One retrieved multiple choice anchor asks for the definition of an antiderivative, with four answer options and a short deterministic solution. This calibrates the output to a low level calculus multiple choice item, but the generated item targets substitution rather than paraphrasing the definition question.

The generated item asks students to compute
\[
\int_{0}^{1}(2x+1)e^{x^2+x}\,dx
\]
using the substitution method. The answer options are: A. $e^2-1$; B. $e-1$; C. $(e^2-1)/2$; D. $e^{x^2+x}\big|_{0}^{1}$.

\textbf{Answer.} A.

\textbf{Solution.} Let $u=x^2+x$, so $du=(2x+1)\,dx$. When $x=0$, $u=0$; when $x=1$, $u=2$. Therefore
\[
\int_{0}^{1}(2x+1)e^{x^2+x}\,dx
=\int_{0}^{2}e^u\,du
=e^u\big|_0^2
=e^2-1.
\]
\end{quote}

\begin{quote}\small
\textbf{True or false cluster example.} The requirement uses the $s=2$ schema in Table \ref{tab:exam_section_schema}. It asks for one common stem, four true or false statements and a detailed solution for all statements. The target topic is geometry, specifically sphere equations in space. One retrieved true or false cluster anchor describes motion in $Oxyz$ and a danger region modeled by a sphere. It supplies the clustered statement structure and intended geometry difficulty. The generated item changes the context to a static sphere equation and asks about center, radius, expansion and inside or on sphere tests.

The generated item gives the sphere
\[
(x-1)^2+(y+2)^2+(z-3)^2=25
\]
and asks whether the following statements are true or false: (a) the center is $(1,-2,3)$ and the radius is 5; (b) if $M(6,-2,3)$ lies on the sphere, then the distance from $M$ to the center is 5; (c) the general equation is $x^2+y^2+z^2-2x+4y-6z+1=0$; and (d) the point $N(2,-1,4)$ lies inside the sphere. The labels are True, True, False and True, respectively.

\textbf{Solution.} From the standard form, the center is $I(1,-2,3)$ and the radius is $R=5$. For $M(6,-2,3)$, $IM=5$, so statement (b) is true. Expanding the sphere gives
\[
x^2+y^2+z^2-2x+4y-6z-11=0,
\]
so statement (c) is false. Finally, $IN=\sqrt{3}<5$, so $N$ lies inside the sphere and statement (d) is true.
\end{quote}

\subsection{Consecutive Verification and Repair}
\label{sec:cvm}
CVM reduces the risk that an invalid question is accepted after a single verification pass. A verifier role $\mathcal{M}_{\mathrm{ver}}$ reviews the generated question by solving it, checking the official answer and solution and verifying that it satisfies the requirement $R$. In the main configuration, the verifier is separated from the generator by different prompts and decoding settings, but it is not a completely different model family.

A question is accepted only if it passes $N=3$ verification checks in a row.

Formally, let $X_t=(P_t,S_t,M_t)$ denote the candidate item tuple at verification step $t$, with initial counter $c_0=0$. At each step, the verifier returns
\begin{equation}
(d_t,E_t)=\mathcal{M}_{\mathrm{ver}}(X_t,R),
\label{eq:cvm_verify}
\end{equation}
where $d_t\in\{\textsc{pass},\textsc{fail}\}$ is the verifier decision and $E_t$ is the failure trace. CVM then updates the consecutive pass counter as
\begin{equation}
c_t=
\begin{cases}
c_{t-1}+1, & d_t=\textsc{pass},\\
0, & d_t=\textsc{fail}.
\end{cases}
\label{eq:cvm_counter}
\end{equation}
If the verifier fails the candidate, the next candidate is repaired using the failure trace:
\begin{equation}
X_{t+1}=
\begin{cases}
X_t, & d_t=\textsc{pass},\\
\mathcal{M}_{\mathrm{gen}}(R,B_R,E_t), & d_t=\textsc{fail}.
\end{cases}
\label{eq:cvm_repair}
\end{equation}
The candidate is accepted only when $c_t=N$ before the repair budget $T$ is exhausted; otherwise, the requirement is returned as a rejected sample. This formulation makes explicit that CVM is a stateful verification and repair process in which any failed check resets the acceptance trajectory.

If a check fails, the verifier returns a failure trace $E$ that explains the error and the reason for rejection. The generator then revises the question using this feedback while keeping the original requirement unchanged. Algorithm \ref{alg:verimath} summarizes the process.

\begin{verimathalgorithm}{VeriMath Inference With Consecutive Verification}
\label{alg:verimath}
\begin{tabular}{@{}r p{0.84\linewidth}@{}}
1 & \textbf{Input:} requirement $R$, corpus $\mathcal{D}_{\mathrm{ref}}$, anchor count $k$, pass threshold $N$, repair limit $T$.\\
2 & Select anchors $B_R \leftarrow \operatorname{SelectAnchors}(R,\mathcal{D}_{\mathrm{ref}},k)$.\\
3 & Generate initial candidate $(P,S,M) \leftarrow \mathcal{M}_{\mathrm{gen}}(R,B_R)$.\\
4 & Self-improve $(P,S,M)$ using the schema associated with $s$.\\
5 & Set $passes \leftarrow 0$ and $repairs \leftarrow 0$.\\
6 & \textbf{while} $passes<N$ and $repairs<T$ \textbf{do}\\
7 & \quad $(d,E) \leftarrow \mathcal{M}_{\mathrm{ver}}(P,S,M,R)$.\\
8 & \quad \textbf{if} $d=\textsc{pass}$ \textbf{then} $passes \leftarrow passes+1$.\\
9 & \quad \textbf{else} revise $(P,S,M) \leftarrow \mathcal{M}_{\mathrm{gen}}(R,B_R,E)$; set $passes \leftarrow 0$ and $repairs \leftarrow repairs+1$.\\
10 & \textbf{end while}.\\
11 & \textbf{if} $passes=N$ \textbf{then return} accepted tuple $(P,S,M)$; otherwise \textbf{return} rejected sample.\\
\end{tabular}
\end{verimathalgorithm}

CVM requires repeated successful verification, making acceptance less dependent on one favorable verifier decision. It also turns failed verification traces into revision feedback, allowing the system to fix unclear statements, incorrect answers, formatting problems and difficulty level mismatches before rejecting a candidate. In all reported experiments, a candidate receives at most three verifier guided repair attempts.

\noindent\textbf{Representative verifier prompt.}
\begin{quote}\small
You are an expert mathematician and a strict mathematics teacher. Verify the provided mathematical problem rigorously. A problem is correct only if every step is justified. Act as a verifier, not as an exam setter. Do not fix errors. Classify issues as incorrect problem or answer, ill-posed or ambiguous problem, or formal/language/formatting issue. End with either \texttt{IS\_CORRECT: TRUE} or \texttt{IS\_CORRECT: FALSE}.
\end{quote}

\noindent\textbf{Representative corrector prompt.}
\begin{quote}\small
Below is a bug report from a verifier. Revise the problem to address it. Preserve the original requirement unless the verifier identifies the requirement interpretation as the source of failure.
\end{quote}

\subsubsection{Worked Example of Verifier-Guided Revision}
\label{sec:cvm-example}
This example uses the $s=3$ short-answer schema from the Vietnamese case-study configuration in Table \ref{tab:exam_section_schema}, with Application level and the learning outcome ``graph transformations.'' Three same-schema and same-level anchors are used only to calibrate format and reasoning depth. The trace in Table \ref{tab:worked_generation_trace} focuses on the generated candidate, verifier feedback and subsequent revisions.

\begin{table}[!t]
\caption{Verifier-Guided Revision Trace for a Generated Item}
\label{tab:worked_generation_trace}
\centering
\scriptsize
\renewcommand{\arraystretch}{1.08}
\begin{tabular}{p{0.13\textwidth}p{0.19\textwidth}p{0.56\textwidth}}
\toprule
\textbf{Step} & \textbf{Decision or output} & \textbf{Relevant content} \\
\midrule
Init & Candidate generated & Stem transforms $y=f(x)=x^3$ by horizontal compression, vertical stretch, reflection over the $x$-axis and a shift right by 1. It asks for the $x$-coordinate where the tangent is horizontal. Answer: 1. \\
Self-improve & Candidate rewritten & The candidate changes to a horizontal stretch by 3, vertical compression by $1/2$, reflection over the $y$-axis and upward shift by 2. It asks for slope $-3$ but reports answer 2. \\
Verify 1 & \texttt{IS\_CORRECT: FALSE} & The verifier derives $h(x)=-x^3/54+2$ and $h'(x)=-x^2/18$. Solving $h'(x)=-3$ gives $x=\pm3\sqrt{6}$, so the official answer 2 is invalid. \\
Correct 1 & Candidate rewritten & The corrected candidate uses horizontal stretch by 2, vertical compression by $1/2$, reflection over the $y$-axis and upward shift by 1. It derives $h(x)=-x^3/16+1$ but reports answer 0.5. \\
Verify 2 & \texttt{IS\_CORRECT: FALSE} & The verifier accepts the transformation and derivative but rejects the answer: $h'(x)=-3$ gives $x=\pm4$, not 0.5. \\
Correct 2 & Candidate rewritten & The candidate reports answer $-4$, but the stem still has no extra condition, so both $x=4$ and $x=-4$ satisfy the slope condition. \\
Verify 3 & \texttt{IS\_CORRECT: FALSE} & The verifier rejects the item as incomplete because the official answer omits the valid solution $x=4$. \\
Correct 3 & Candidate rewritten & The stem adds the condition $x<0$, preserving the same transformation and slope requirement. The final answer becomes $-4$. \\
Verify 4 & \texttt{IS\_CORRECT: TRUE} & First successful verification pass. The verifier checks the graph transformations, derivative and condition $x<0$. \\
Verify 5 & \texttt{IS\_CORRECT: TRUE} & Second consecutive successful verification pass. \\
Verify 6 & \texttt{IS\_CORRECT: TRUE} & Third consecutive successful verification pass. The item is accepted by CVM. \\
\bottomrule
\end{tabular}
\end{table}

The final accepted item can be rendered in English as follows.
\begin{quote}\small
\textbf{Stem.} The graph of $y=h(x)$ is obtained from the graph of $y=f(x)=x^3$ by applying a horizontal stretch by factor 2, a vertical compression by factor $1/2$, reflection over the $y$-axis and an upward shift by 1 unit. Determine the $x$-coordinate of the point where the tangent to the graph of $h(x)$ has slope $-3$ and $x<0$.

\textbf{Answer.} $-4$.

\textbf{Solution outline.} Since
\[
h(x)=\frac{1}{2}f\left(-\frac{x}{2}\right)+1
=-\frac{x^3}{16}+1,
\]
we have $h'(x)=-3x^2/16$. Setting $h'(x)=-3$ gives $x^2=16$, so $x=\pm4$. The condition $x<0$ selects $x=-4$.
\end{quote}

\section{Experiments}
\label{sec:experiments}
The experiments instantiate VeriMath in the Vietnamese Grade 12 setting and address two questions aligned with its intended use as an assessment-authoring support tool. First, can the end-to-end framework produce structurally complete, novel, verifier-accepted drafts under the curriculum and format constraints of this case study? Second, do human teachers judge those drafts as ready for educational use? The first question is answered through the diagnostics in Section \ref{sec:pipeline-diagnostics}. The second is answered through the expert review and failure analysis in Sections \ref{sec:human-readiness} and \ref{sec:failure-analysis}.

\subsection{Experimental Setup}
\label{sec:experimental-setup}
\subsubsection{Models and Decoding}
The generation pipeline is evaluated with two backbones: \texttt{Qwen3-8B} and \texttt{gpt-oss-20b}. Within each backbone run, the same underlying model is used for generation, self-improvement, verification and correction, but each step uses a different role prompt. Thus, the verifier is a separately prompted role rather than an independent model family. Question generation uses random sampling with temperature 1.0. CVM requires three consecutive verifier approvals and allows up to three verifier-guided correction attempts before rejection. After verification, a separate \LaTeX{} formatting pass is applied with up to three repair attempts. Novelty is measured by comparing each generated question with all 4,555 questions in the reference dataset using a multilingual sentence-embedding model.

An automated LLM-judge route was also explored as a separate diagnostic role, but it is not used as primary readiness evidence in this study. The judge prompt asks the model to evaluate exactly four dimensions on a $[0,1]$ scale: requirement alignment, level fidelity, language fluency and mathematical accuracy. It explicitly defines level fidelity as matching the requested teacher-annotated cognitive/difficulty level and requests valid JavaScript Object Notation (JSON) with those four keys and a brief feedback field.

\subsubsection{Data and Splits}
The teacher annotated reference corpus contains 4,555 questions and is divided by exam source into an anchor set and a separate evaluation set. This split helps reduce overlap from highly similar practice questions. Experimental requirements are created independently from the Grade 12 curriculum. Each backbone is evaluated on the same 50 requirements with the schema ratio defined in Section \ref{sec:requirement-set}. After expanding the four statements in each true or false cluster from Table \ref{tab:exam_section_schema}, the requirement set contains 80 learning-outcome slots: 67 from calculus/statistics and 13 from geometry. The historical dataset is used only for anchoring and novelty measurement, not for training the backbone models.

\subsubsection{Scope of Empirical Claims}
This study reports end-to-end full-pipeline runs and completed expert-review records. It does not report controlled baseline or ablation experiments. Therefore, the acceptance rates should not be interpreted as causal evidence for the independent contribution of Difficulty Anchoring, self-improvement, CVM, or formatting repair. The component-level claims in this paper are limited to their designed roles and to qualitative traces of verifier-guided revision. Claims about educational readiness are based on blind expert review rather than on automated verifier decisions. All reported empirical evidence pertains to the Vietnamese Grade 12 instantiation and does not establish performance transfer across curricula or countries.

\subsubsection{Metrics}
The quantitative diagnostics report accepted outputs, embedding-based novelty, verifier-call counts, verifier-guided repair calls, runtime and token usage. Novelty is calculated as
\begin{equation}
\operatorname{Nov}(P)=1-\max_{b\in\mathcal{D}_{\mathrm{ref}}}
\frac{\mathbf{v}_{P}\cdot\mathbf{v}_{b}}
{\|\mathbf{v}_{P}\|\|\mathbf{v}_{b}\|}.
\label{eq:novelty}
\end{equation}
where $\mathbf{v}_{P}$ and $\mathbf{v}_{b}$ are sentence embeddings of the generated problem and reference item $b$, respectively. These operational metrics are interpreted as production and review-cost indicators. The primary item-quality measure is the expert-readiness decision defined by the four-criterion teacher rubric.

\subsection{Pipeline Diagnostics}
\label{sec:pipeline-diagnostics}
\begin{figure*}[!t]
\centering

\definecolor{vmBlue}{HTML}{0072B2}
\definecolor{vmAmber}{HTML}{D55E00}
\definecolor{vmBorder}{HTML}{374151}
\definecolor{vmGrid}{HTML}{E5E7EB}
\definecolor{vmText}{HTML}{111827}
\definecolor{vmPair}{HTML}{9CA3AF}

\pgfplotsset{
  vm cost axis/.style={
    scale only axis,
    axis x line*=bottom,
    axis y line*=left,
    axis line style={vmBorder!55, line width=0.35pt},
    tick style={vmBorder!55, line width=0.3pt},
    tick align=outside,
    xmajorgrids=true,
    grid style={solid, vmGrid, line width=0.25pt},
    scaled x ticks=false,
    xticklabel={\pgfmathprintnumber[fixed,precision=0,1000 sep={,}]{\tick}},
    tick label style={font=\footnotesize, text=vmText},
    x tick label style={font=\footnotesize},
    label style={font=\small, text=vmText},
    title style={font=\small\bfseries, text=vmText, at={(0,1.03)}, anchor=south west},
    y tick label style={font=\footnotesize, text=vmText, align=right},
    every node near coord/.append style={font=\footnotesize, text=vmText, inner sep=1pt},
    clip=false,
  },
  qwen dot/.style={only marks, mark=*, mark size=1.8pt, draw=vmBlue!70!black, fill=vmBlue},
  oss dot/.style={only marks, mark=square*, mark size=1.65pt, draw=vmAmber!70!black, fill=vmAmber},
}

\begin{tikzpicture}
\begin{axis}[
  hide axis,
  xmin=0,xmax=1,ymin=0,ymax=1,
  legend columns=2,
  legend style={draw=none, fill=none, font=\footnotesize,
    /tikz/every even column/.append style={column sep=1em}},
]
\addlegendimage{qwen dot}
\addlegendentry{Qwen/Qwen3-8B}
\addlegendimage{oss dot}
\addlegendentry{gpt-oss-20b}
\end{axis}
\end{tikzpicture}
\vspace{-0.3em}

\begin{minipage}[t]{0.96\textwidth}
\centering
\begin{tikzpicture}
\begin{axis}[
  vm cost axis,
  width=0.74\linewidth,
  height=5.6cm,
  xmin=0, xmax=10500,
  xtick={0,2500,5000,7500,10000},
  xlabel={Tokens per accepted item},
  ymin=0.55, ymax=5.45,
  ytick={1,2,3,4,5},
  yticklabels={LaTeX repair,Correct,Verify,Self-improve,Initial gen.},
  title={(a) Token cost},
  nodes near coords={\pgfmathprintnumber[fixed,precision=0,1000 sep={,}]{\pgfplotspointmeta}},
  point meta=x,
]
\draw[vmPair, line width=0.45pt] (axis cs:759,1) -- (axis cs:1510,1);
\draw[vmPair, line width=0.45pt] (axis cs:3382,2) -- (axis cs:2363,2);
\draw[vmPair, line width=0.45pt] (axis cs:8280,3) -- (axis cs:9500,3);
\draw[vmPair, line width=0.45pt] (axis cs:4560,4) -- (axis cs:5531,4);
\draw[vmPair, line width=0.45pt] (axis cs:4678,5) -- (axis cs:6108,5);

\addplot+[qwen dot, every node near coord/.append style={anchor=south, yshift=2pt}]
coordinates {
  (759,1.08)
  (3382,2.08)
  (8280,3.08)
  (4560,4.08)
  (4678,5.08)
};

\addplot+[oss dot, every node near coord/.append style={anchor=north, yshift=-2pt}]
coordinates {
  (1510,0.92)
  (2363,1.92)
  (9500,2.92)
  (5531,3.92)
  (6108,4.92)
};
\end{axis}
\end{tikzpicture}
\end{minipage}
\par\medskip
\begin{minipage}[t]{0.96\textwidth}
\centering
\begin{tikzpicture}
\begin{axis}[
  vm cost axis,
  width=0.74\linewidth,
  height=5.6cm,
  xmin=0, xmax=5300,
  xtick={0,1000,2000,3000,4000,5000},
  xlabel={Recorded seconds},
  ymin=0.55, ymax=5.45,
  ytick={1,2,3,4,5},
  yticklabels={LaTeX repair,Correct,Verify,Self-improve,Initial gen.},
  title={(b) Recorded duration},
  nodes near coords={\pgfmathprintnumber[fixed,precision=0,1000 sep={,}]{\pgfplotspointmeta}},
  point meta=x,
]
\draw[vmPair, line width=0.45pt] (axis cs:272.2,1) -- (axis cs:978.1,1);
\draw[vmPair, line width=0.45pt] (axis cs:1250.8,2) -- (axis cs:1008.0,2);
\draw[vmPair, line width=0.45pt] (axis cs:2157.9,3) -- (axis cs:2688.7,3);
\draw[vmPair, line width=0.45pt] (axis cs:1033.8,4) -- (axis cs:2748.2,4);
\draw[vmPair, line width=0.45pt] (axis cs:1267.1,5) -- (axis cs:5036.8,5);

\addplot+[qwen dot, every node near coord/.append style={anchor=south, yshift=2pt}]
coordinates {
  (272.2,1.08)
  (1250.8,2.08)
  (2157.9,3.08)
  (1033.8,4.08)
  (1267.1,5.08)
};

\addplot+[oss dot, every node near coord/.append style={anchor=north, yshift=-2pt}]
coordinates {
  (978.1,0.92)
  (1008.0,1.92)
  (2688.7,2.92)
  (2748.2,3.92)
  (5036.8,4.92)
};
\end{axis}
\end{tikzpicture}
\end{minipage}

\caption{Pipeline cost by processing step for accepted outputs. Panel (a) reports token cost normalized per accepted item; Panel (b) reports aggregate recorded duration. Paired points within a row compare the two models}
\label{fig:pipeline_cost_breakdown}
\end{figure*}
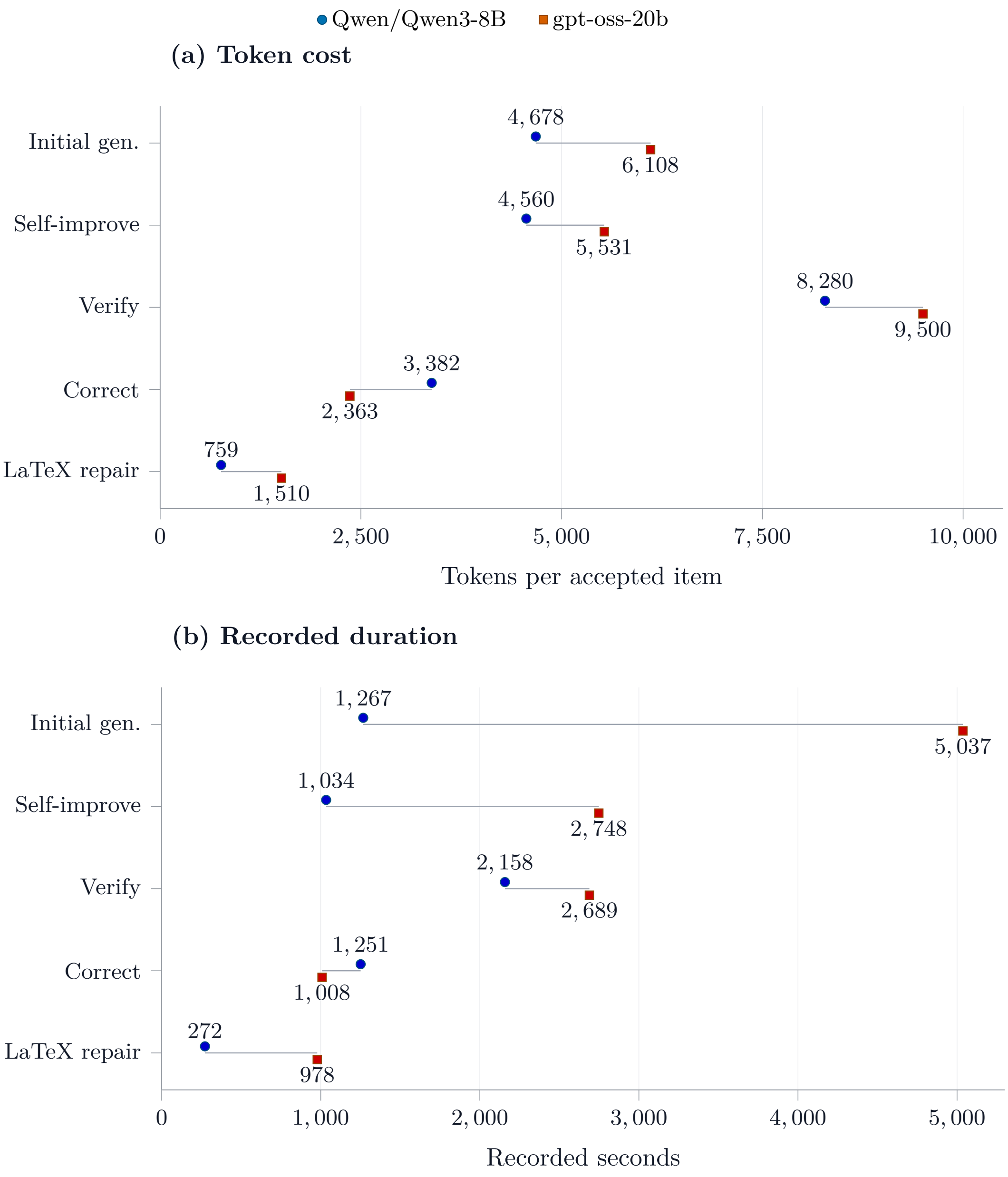

Table \ref{tab:pipeline_diagnostics} reports full-pipeline diagnostics for the two backbones. Accepted outputs are items that complete generation, self-improvement, CVM verification and final formatting. Novelty is averaged over accepted outputs. Verifier and repair calls are normalized by accepted output to make the inference cost of CVM visible. Tables \ref{tab:runtime_token_cost}, \ref{tab:section_cost} and \ref{tab:step_cost} report runtime and token usage from the same traces. Full item runtime is wall-clock time for a generated item, while step duration is the duration recorded for individual LLM calls. Fig. \ref{fig:pipeline_cost_breakdown} summarizes the step-level token and duration profile.

\begin{table}[!t]
\caption{Full-Pipeline Diagnostics for the Reported Backbones}
\label{tab:pipeline_diagnostics}
\centering
\scriptsize
\setlength{\tabcolsep}{2.2pt}
\renewcommand{\arraystretch}{1.12}
\begin{tabular}{lccccc}
\toprule
\textbf{Backbone} & \textbf{Accepted} & \textbf{Acceptance} & \textbf{Novelty $\uparrow$} & \textbf{\shortstack{Verifier\\calls/item}} & \textbf{Repairs/item} \\
\midrule
\texttt{Qwen3-8B} & 47/50 & 94.0\% & 0.813 & 3.85 & 0.70 \\
\texttt{gpt-oss-20b} & 47/50 & 94.0\% & 0.787 & 3.53 & 0.34 \\
\bottomrule
\end{tabular}
\end{table}

\begin{table}[!t]
\caption{Runtime and Token Cost of the Full Pipeline}
\label{tab:runtime_token_cost}
\centering
\scriptsize
\setlength{\tabcolsep}{2pt}
\renewcommand{\arraystretch}{1.12}
\begin{tabular}{p{0.22\linewidth}cccccc}
\toprule
\textbf{Backbone} & \textbf{\shortstack{Mean\\time/item}} & \textbf{\shortstack{Median\\time/item}} & \textbf{\shortstack{Input\\tok./item}} & \textbf{\shortstack{Output\\tok./item}} & \textbf{\shortstack{Total\\tok./item}} & \textbf{\shortstack{Recorded\\tok.}} \\
\midrule
\texttt{Qwen3-8B} & 136.1 s & 91.4 s & 10,547 & 11,111 & 21,659 & 1,125,879 \\
\texttt{gpt-oss-20b} & 468.0 s & 318.7 s & 11,425 & 13,588 & 25,012 & 1,175,581 \\
\bottomrule
\end{tabular}
\end{table}

\begin{table}[!t]
\caption{Runtime and Token Cost by Examination Schema}
\label{tab:section_cost}
\centering
\scriptsize
\setlength{\tabcolsep}{1.8pt}
\renewcommand{\arraystretch}{1.12}
\begin{tabular}{p{0.15\linewidth}lcccccc}
\toprule
\textbf{Backbone} & \textbf{Schema} & \textbf{\shortstack{Accepted/\\recorded}} & \textbf{\shortstack{Mean\\time}} & \textbf{\shortstack{Median\\time}} & \textbf{\shortstack{Input\\tok.}} & \textbf{\shortstack{Output\\tok.}} & \textbf{\shortstack{Total\\tok.}} \\
\midrule
\texttt{Qwen3-8B} & $s=1$ & 29/30 & 125.8 s & 83.4 s & 9,372 & 10,625 & 19,997 \\
\texttt{Qwen3-8B} & $s=2$ & 8/9 & 162.3 s & 105.2 s & 15,175 & 11,085 & 26,260 \\
\texttt{Qwen3-8B} & $s=3$ & 10/10 & 145.3 s & 130.1 s & 10,255 & 12,542 & 22,798 \\
\texttt{gpt-oss-20b} & $s=1$ & 30/30 & 367.3 s & 309.3 s & 9,744 & 12,541 & 22,285 \\
\texttt{gpt-oss-20b} & $s=2$ & 7/10 & 941.2 s & 659.0 s & 19,265 & 19,304 & 38,569 \\
\texttt{gpt-oss-20b} & $s=3$ & 10/10 & 438.9 s & 290.2 s & 10,979 & 12,727 & 23,705 \\
\bottomrule
\end{tabular}
\end{table}

\begin{table}[!t]
\caption{Token and Duration Breakdown by Pipeline Step}
\label{tab:step_cost}
\centering
\scriptsize
\setlength{\tabcolsep}{1.8pt}
\renewcommand{\arraystretch}{1.12}
\begin{tabular}{p{0.15\linewidth} p{0.22\linewidth} ccccc}
\toprule
\textbf{Backbone} & \textbf{Step} & \textbf{Calls} & \textbf{\shortstack{Mean\\sec./call}} & \textbf{\shortstack{Total\\sec.}} & \textbf{\shortstack{Mean\\tok./call}} & \textbf{Tok./item} \\
\midrule
\texttt{Qwen3-8B} & Initial generation & 47 & 27.0 & 1,267.1 & 4,678 & 4,678 \\
\texttt{Qwen3-8B} & Self-improvement & 47 & 22.0 & 1,033.8 & 4,560 & 4,560 \\
\texttt{Qwen3-8B} & Verification & 171 & 12.6 & 2,157.9 & 2,276 & 8,280 \\
\texttt{Qwen3-8B} & Verifier-guided correction & 27 & 46.3 & 1,250.8 & 5,887 & 3,382 \\
\texttt{Qwen3-8B} & \LaTeX{} formatting repair & 7 & 38.9 & 272.2 & 5,093 & 759 \\
\texttt{gpt-oss-20b} & Initial generation & 47 & 107.2 & 5,036.8 & 6,108 & 6,108 \\
\texttt{gpt-oss-20b} & Self-improvement & 47 & 58.5 & 2,748.2 & 5,531 & 5,531 \\
\texttt{gpt-oss-20b} & Verification & 166 & 16.2 & 2,688.7 & 2,690 & 9,500 \\
\texttt{gpt-oss-20b} & Verifier-guided correction & 16 & 63.0 & 1,008.0 & 6,941 & 2,363 \\
\texttt{gpt-oss-20b} & \LaTeX{} formatting repair & 10 & 97.8 & 978.1 & 7,097 & 1,510 \\
\bottomrule
\end{tabular}
\end{table}

The operational results show that both backbones can complete most curriculum-constrained requirements under the full framework. However, the numbers in Table \ref{tab:pipeline_diagnostics} should be read as draft-production diagnostics, not as teacher-readiness scores. Table \ref{tab:section_cost} follows the schema notation from Table \ref{tab:exam_section_schema}; the recorded denominator denotes items with complete runtime and token traces for that schema. The cost tables also show that repeated verification is a substantial part of the authoring workflow. This cost may be acceptable for producing reviewable item drafts, but it is too high for uncontrolled mass generation without a teacher-facing selection interface.

\subsection{Human Readiness Evaluation}
\label{sec:human-readiness}
The expert evaluation follows the blind protocol described in Section \ref{sec:reference-corpus}. Mathematics educators review generated items without seeing the generation source, model identity or process traces. They rate each item on four criteria. Mathematical correctness and answer consistency are recorded as one binary label, and alignment with the assigned learning outcome is recorded as another binary label. Match to the required cognitive or difficulty level and clarity or readiness for examination use are scored on a 1 to 5 scale.

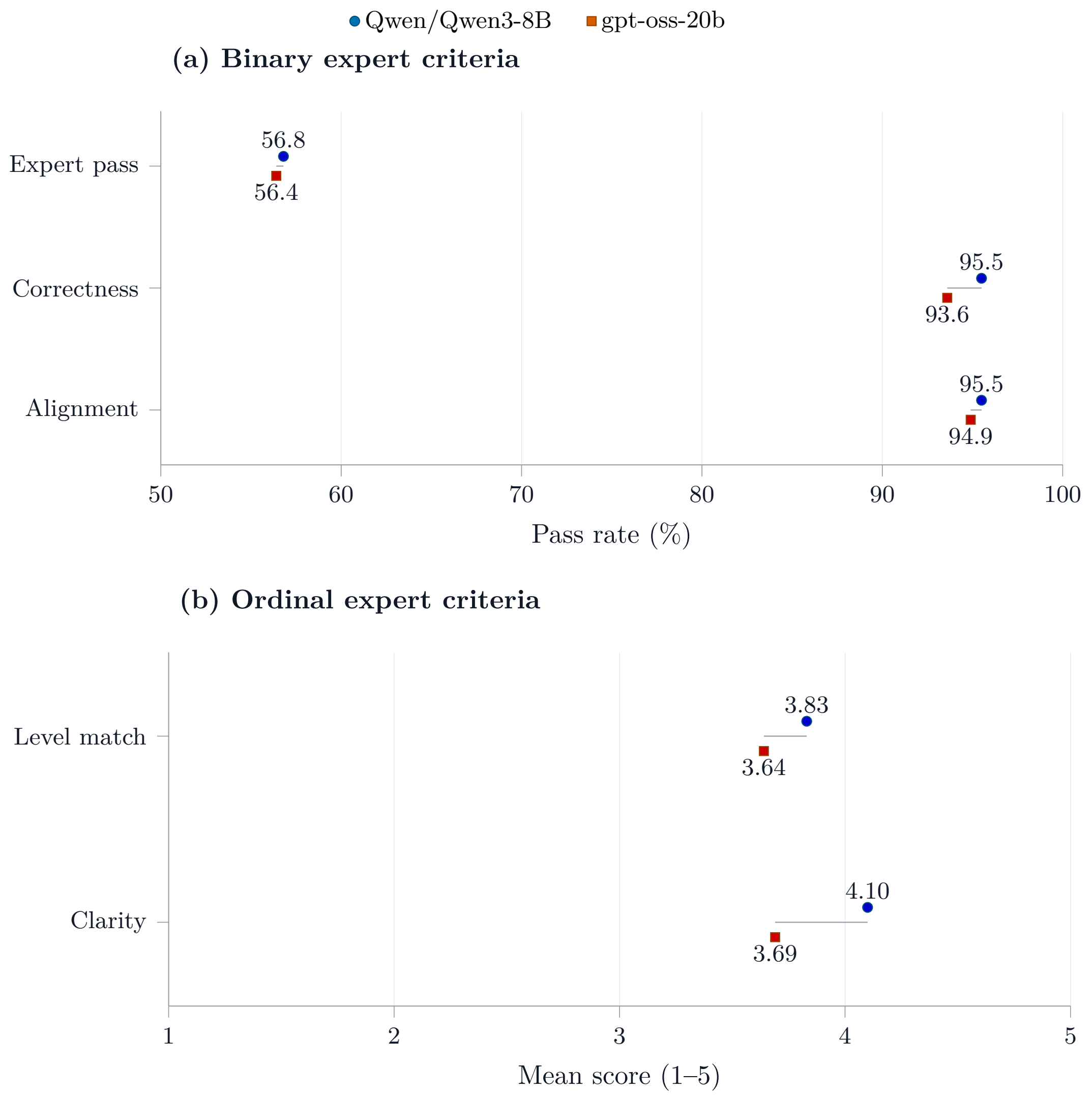
\begin{figure*}[!t]
\centering

\definecolor{vmBlue}{HTML}{0072B2}
\definecolor{vmAmber}{HTML}{D55E00}
\definecolor{vmBorder}{HTML}{374151}
\definecolor{vmGrid}{HTML}{E5E7EB}
\definecolor{vmText}{HTML}{111827}
\definecolor{vmPair}{HTML}{9CA3AF}

\pgfplotsset{
  vm dot axis/.style={
    scale only axis,
    axis x line*=bottom,
    axis y line*=left,
    axis line style={vmBorder!55, line width=0.35pt},
    tick style={vmBorder!55, line width=0.3pt},
    tick align=outside,
    xmajorgrids=true,
    grid style={solid, vmGrid, line width=0.25pt},
    tick label style={font=\footnotesize, text=vmText},
    label style={font=\small, text=vmText},
    title style={font=\small\bfseries, text=vmText, at={(0,1.03)}, anchor=south west},
    y tick label style={font=\footnotesize, text=vmText, align=right},
    every node near coord/.append style={font=\footnotesize, text=vmText, inner sep=1pt},
    clip=false,
  },
  qwen dot/.style={only marks, mark=*, mark size=1.8pt, draw=vmBlue!70!black, fill=vmBlue},
  oss dot/.style={only marks, mark=square*, mark size=1.65pt, draw=vmAmber!70!black, fill=vmAmber},
}

\begin{tikzpicture}
\begin{axis}[
  hide axis,
  xmin=0,xmax=1,ymin=0,ymax=1,
  legend columns=2,
  legend style={draw=none, fill=none, font=\footnotesize,
    /tikz/every even column/.append style={column sep=1em}},
]
\addlegendimage{qwen dot}
\addlegendentry{Qwen/Qwen3-8B}
\addlegendimage{oss dot}
\addlegendentry{gpt-oss-20b}
\end{axis}
\end{tikzpicture}
\vspace{-0.3em}

\begin{minipage}[t]{0.96\textwidth}
\centering
\begin{tikzpicture}
\begin{axis}[
  vm dot axis,
  width=0.74\linewidth,
  height=4.6cm,
  xmin=50, xmax=100,
  xtick={50,60,70,80,90,100},
  xlabel={Pass rate (\%)},
  ymin=0.55, ymax=3.45,
  ytick={1,2,3},
  yticklabels={Alignment,Correctness,Expert pass},
  title={(a) Binary expert criteria},
  nodes near coords={\pgfmathprintnumber[fixed,precision=1]{\pgfplotspointmeta}},
  point meta=x,
]
\draw[vmPair, line width=0.45pt] (axis cs:56.8,3) -- (axis cs:56.4,3);
\draw[vmPair, line width=0.45pt] (axis cs:95.5,2) -- (axis cs:93.6,2);
\draw[vmPair, line width=0.45pt] (axis cs:95.5,1) -- (axis cs:94.9,1);

\addplot+[qwen dot, every node near coord/.append style={anchor=south, yshift=2pt}]
coordinates {
  (95.5,1.08)
  (95.5,2.08)
  (56.8,3.08)
};

\addplot+[oss dot, every node near coord/.append style={anchor=north, yshift=-2pt}]
coordinates {
  (94.9,0.92)
  (93.6,1.92)
  (56.4,2.92)
};
\end{axis}
\end{tikzpicture}
\end{minipage}
\par\medskip
\begin{minipage}[t]{0.96\textwidth}
\centering
\begin{tikzpicture}
\begin{axis}[
  vm dot axis,
  width=0.74\linewidth,
  height=4.6cm,
  xmin=1, xmax=5,
  xtick={1,2,3,4,5},
  xlabel={Mean score (1--5)},
  ymin=0.55, ymax=2.45,
  ytick={1,2},
  yticklabels={Clarity,Level match},
  title={(b) Ordinal expert criteria},
  nodes near coords={\pgfmathprintnumber[fixed,fixed zerofill,precision=2]{\pgfplotspointmeta}},
  point meta=x,
]
\draw[vmPair, line width=0.45pt] (axis cs:4.10,1) -- (axis cs:3.69,1);
\draw[vmPair, line width=0.45pt] (axis cs:3.83,2) -- (axis cs:3.64,2);

\addplot+[qwen dot, every node near coord/.append style={anchor=south, yshift=2pt}]
coordinates {
  (4.10,1.08)
  (3.83,2.08)
};

\addplot+[oss dot, every node near coord/.append style={anchor=north, yshift=-2pt}]
coordinates {
  (3.69,0.92)
  (3.64,1.92)
};
\end{axis}
\end{tikzpicture}
\end{minipage}

\caption{Expert-review outcomes for Qwen/Qwen3-8B and gpt-oss-20b. Binary criteria are shown as pass rates; ordinal criteria are shown as mean ratings on a 1--5 scale. Paired points within a row compare the two models}
\label{fig:expert_review_profile}
\end{figure*}

A generated item passes expert review when correctness and alignment both pass and the level match and clarity scores are both at least 4. When multiple ratings are available, the reported score is the average of independent ratings. Reviewers also mark error flags for wrong answer, invalid problem statement, multiple valid answers, missing or incorrect solution step, wrong learning outcome, too easy, too hard, ambiguous or missing condition and formatting or formula error. Fig. \ref{fig:expert_review_profile} separates binary criteria from 1 to 5 rating criteria.

The intended review pool is the 50-item requirement set for each reported generation backbone. The quantitative analysis below uses completed expert-review records: 44 items for \texttt{Qwen3-8B} and 39 items for \texttt{gpt-oss-20b}. Each reviewer sees the problem statement, official answer, solution, requested schema value $s$, target curriculum learning outcome and requested cognitive or difficulty level. A reference question from the historical corpus may be shown for difficulty, format and novelty calibration.

\begin{table}[!t]
\caption{Blind Expert Review Under the Four-Criterion Teacher Rubric}
\label{tab:blind_expert_review}
\centering
\scriptsize
\setlength{\tabcolsep}{2pt}
\renewcommand{\arraystretch}{1.12}
\begin{tabular}{p{0.22\linewidth}cccccc}
\toprule
\textbf{Model} & \textbf{\shortstack{Reviewed\\items}} & \textbf{\shortstack{Expert\\pass $\uparrow$}} & \textbf{Correct. $\uparrow$} & \textbf{Align. $\uparrow$} & \textbf{Level $\uparrow$} & \textbf{Clarity $\uparrow$} \\
\midrule
\texttt{Qwen3-8B} & 44 & 56.8\% & 95.5\% & 95.5\% & 3.83 & 4.10 \\
\texttt{gpt-oss-20b} & 39 & 56.4\% & 93.6\% & 94.9\% & 3.64 & 3.69 \\
\bottomrule
\end{tabular}
\end{table}

Table \ref{tab:blind_expert_review} is the central quality result. Correctness and curriculum alignment are high, but expert pass rates are much lower because readiness also requires level fit and clear exam-style presentation. The difference between the 94.0\% verifier-accepted rate and the approximately 56\% expert-pass rate shows that CVM is useful as a screening and repair mechanism, but automatic verification is not sufficient for final educational use. VeriMath should therefore be deployed as a draft authoring aid with a human quality gate, not as an autonomous examination generator.

The expert labels also provide a human reference for future judge calibration. In this paper, they are reported directly rather than converted into automated-judge agreement claims.

\subsection{Novelty and Leakage Checks}
\label{sec:novelty-leakage}
Embedding-based novelty may fail to detect reused templates, especially when a generated item changes only numerical values while keeping the same story and solution method as an anchor. The present paper therefore reports embedding novelty only as a pipeline diagnostic. We do not treat the novelty scores as a complete leakage audit. For assessment authoring, a teacher-facing system should also support nearest-neighbor inspection, symbolic-template comparison and manual review of any generated item whose solution structure is close to a historical question.

\subsection{Expert Failure Analysis and Design Implications}
\label{sec:failure-analysis}
Failure analysis uses the error flags recorded in the expert rubric. Table \ref{tab:expert_error_flags} summarizes the flags observed in reviewed items.

\begin{table}[!t]
\caption{Expert-Marked Error Flags in Reviewed Generated Items}
\label{tab:expert_error_flags}
\centering
\small
\renewcommand{\arraystretch}{1.12}
\begin{tabular}{lcc}
\toprule
\textbf{Error flag} & \textbf{Count} & \textbf{Share} \\
\midrule
Invalid problem statement & 14 & 43.8\% \\
Formatting or formula error & 5 & 15.6\% \\
Missing or incorrect solution step & 3 & 9.4\% \\
Wrong learning outcome & 2 & 6.2\% \\
Too hard & 2 & 6.2\% \\
Ambiguous or missing condition & 2 & 6.2\% \\
Wrong answer & 2 & 6.2\% \\
Multiple valid answers & 1 & 3.1\% \\
Too easy & 1 & 3.1\% \\
\bottomrule
\end{tabular}
\end{table}

The counts in Table \ref{tab:expert_error_flags} sum to 32 marked flags, and shares are calculated over this flag total rather than over reviewed items. The failure profile gives three design implications for assessment-authoring technology. First, invalid problem statements are more frequent than explicit wrong-answer flags, so future user interfaces should help teachers inspect assumptions, domains and uniqueness conditions before considering an item ready. Second, formatting and formula errors remain visible to teachers even after a \LaTeX{} repair pass, suggesting that mathematical rendering checks should be integrated with human preview. Third, wrong learning outcome and difficulty-mismatch flags show that solver-style verification cannot fully judge curriculum appropriateness; teacher review is needed to decide whether an item tests the intended concept at the intended level.

\subsubsection{Examples From Expert-Reviewed Items}
The examples below are English renderings of reviewed Vietnamese items. They show how the rubric separates mathematical validity, curriculum alignment, difficulty match and clarity.

\begin{quote}\small
\textbf{\texttt{gpt-oss-20b} pass example.} This item uses the $s=1$ multiple choice schema in Table \ref{tab:exam_section_schema} and targets basic exponential and power-function integration at Recognition level. The generated item gives $f'(x)=3x^2+4e^x$ and $f(0)=1$, then asks students to compute $\int_0^2 f(x)\,dx$. The answer options are A. $4e^2-6$; B. $4e^2-2$; C. $4e^2+6$; D. $4e^2+2$.

The official answer is A. Integrating $f'(x)$ gives $f(x)=x^3+4e^x+C$. Since $f(0)=1$, $C=-3$, so $f(x)=x^3+4e^x-3$. Therefore
\[
\int_0^2 f(x)\,dx
=\left[\frac{x^4}{4}+4e^x-3x\right]_0^2
=4e^2-6.
\]
Two reviewers rated the item as correct and aligned, with level score 4.0 and clarity score 4.0; the expert decision is pass.
\end{quote}

\begin{quote}\small
\textbf{\texttt{gpt-oss-20b} fail example.} This item uses the $s=1$ multiple choice schema in Table \ref{tab:exam_section_schema} and targets the general equation of a plane in coordinate geometry at Comprehension level. The generated item asks for the plane $P$ passing through $M(1,-2,3)$ and containing the line
\[
\frac{x}{2}=\frac{y}{-1}=\frac{z}{4}.
\]
The answer options include A. $5x-2y-3z=0$, B. $5x+2y+3z=0$, C. $-5x+2y+3z=0$ and D. $5x-2y+3z=0$, with A marked as the official answer.

The solution uses direction vector $\vec{u}=(2,-1,4)$, point $P_0=(0,0,0)$ and $\vec{v}=\overrightarrow{MP_0}=(-1,2,-3)$, then computes $\vec{n}=\vec{u}\times\vec{v}=(-5,2,3)$. Using point $M$ gives $-5(x-1)+2(y+2)+3(z-3)=0$, or $5x-2y-3z=0$.

Two reviewers rated the item as mathematically correct and aligned, with level score 4.5, but clarity score 2.0; the expert decision is fail. Marked issues include formatting/formula error and missing or incorrect solution step. One reviewer noted that the vector-product presentation does not match standard high-school curriculum treatment.
\end{quote}

\begin{quote}\small
\textbf{\texttt{Qwen3-8B} pass example.} This item uses the $s=1$ multiple choice schema in Table \ref{tab:exam_section_schema} and targets finding the maximum or minimum of a function on a closed interval at Recognition level. The generated item asks for the maximum value of $f(x)=x^3-3x^2+4$ on $[0,3]$, with answer options A. 4; B. 0; C. $-2$; D. 1.

The official answer is A. Since $f'(x)=3x^2-6x$, the critical points in the interval are $x=0$ and $x=2$. Evaluating the function at $x=0,2,3$ gives $f(0)=4$, $f(2)=0$ and $f(3)=4$. The maximum value is 4. One reviewer rated the item as correct and aligned, with level score 5.0 and clarity score 5.0; the expert decision is pass.
\end{quote}

\section{Discussion}
\label{sec:discussion}

\begin{figure}[!t]
\centering
\begin{minipage}{0.48\textwidth}
    \centering
    \includegraphics[width=\linewidth]{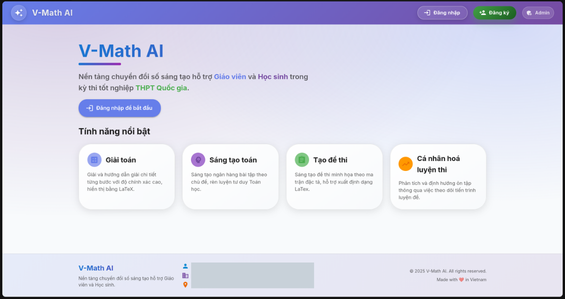}
    \textbf{(a)} Welcome page
\end{minipage}
\hfill
\begin{minipage}{0.48\textwidth}
    \centering
    \includegraphics[width=\linewidth]{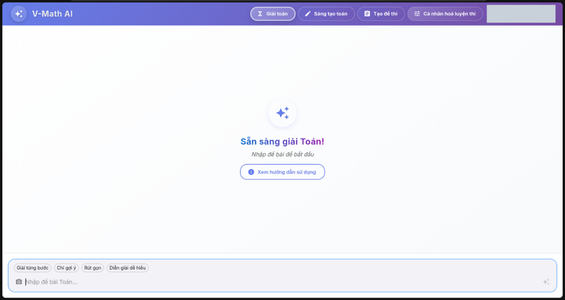}
    \textbf{(b)} Problem solving page
\end{minipage}

\vspace{0.3cm}

\begin{minipage}{0.48\textwidth}
    \centering
    \includegraphics[width=\linewidth]{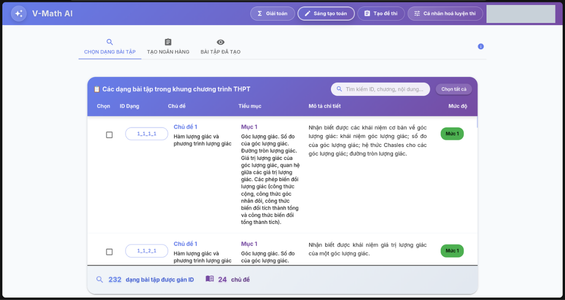}
    \textbf{(c)} Problem generation page
\end{minipage}
\hfill
\begin{minipage}{0.48\textwidth}
    \centering
    \includegraphics[width=\linewidth]{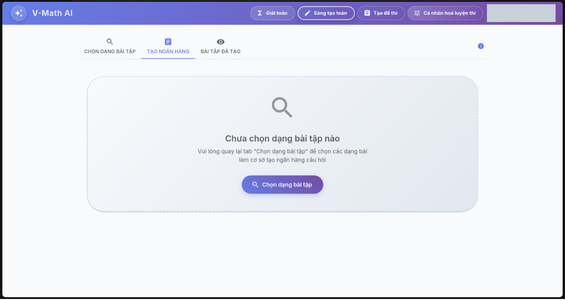}
    \textbf{(d)} Exam synthesis page
\end{minipage}

\caption{Interfaces of the Vietnamese case-study digital mathematics learning platform. The platform includes modules for problem solving, item generation and exam synthesis, while the present study evaluates the curriculum-constrained item-authoring component}
\label{fig:platform_interfaces}
\end{figure}

\subsection{Position Within a Digital Transformation Platform}
\label{sec:discussion-platform}
VeriMath can serve as a curriculum-constrained item-authoring and verification layer within broader digital assessment environments. In such environments, generated drafts can feed teacher-managed workflows for question-bank construction, test assembly, delivery, scoring and revision. This role follows the broader educational principle that generative artificial intelligence should remain connected to pedagogical grounding and teacher judgment \citep{kasneci2023chatgpt,kibar2026instructional}.

In the present Vietnamese implementation, this layer is integrated into a broader web platform that supports teachers and students preparing for the national high school graduation examination in mathematics. The long-term goal of this implementation is a connected educational environment in which teachers can create assessment items, assemble tests, deploy examinations to their classes, collect student responses, apply automatic grading and inspect critiques of generated items and student solutions. Such an environment forms a closed workflow from item authoring, test delivery and automatic scoring to diagnosis, feedback and teacher revision. Figure \ref{fig:platform_interfaces} illustrates the platform modules. This implementation is an example of deployment in one national context; the present study evaluates the item-authoring component rather than the platform as a whole.

\subsection{Strengths and Validation Evidence}
\label{sec:discussion-strengths}
A main strength of the study is that it formulates generation as curriculum constrained assessment authoring rather than open ended text production. Each draft is required to satisfy a specified item format, curriculum learning outcome, cognitive or difficulty level, mathematical answer, solution explanation, novelty condition and teacher review requirement. This framing is important because mathematics items are pedagogical and assessment artifacts, not only fluent linguistic outputs \citep{verschaffel2020word}. This design focus is consistent with the instructional-alignment concerns identified by \citet{kibar2026instructional}.

A second strength is the explicit separation between reusable authoring mechanisms and local adaptation resources. By design, what changes across educational systems is not the anchoring, verification and repair architecture itself, but the curriculum metadata, assessment schemas, difficulty conventions, reference corpus and expert rubric supplied to it. In the Vietnamese Grade 12 case study, the 4,555 item teacher annotated corpus provides examples of notation, topic distribution, assessment schemas and difficulty conventions. Difficulty Anchoring uses this corpus to condition generation on nearby items without treating the anchors as templates to copy. CVM then requires repeated successful verification and uses failure traces for repair before a draft is accepted. This design makes the pipeline more auditable than a single prompt based generator because each accepted item can be linked to its requirement, retrieved anchors, verifier decisions and repair history. The case study demonstrates this adaptation mechanism in one context; it does not establish equivalent performance in other educational systems.

The validation within this case study is deliberately multi layer. The same 50 requirements are tested with two model backbones, and both runs produce verifier accepted drafts for 47 requirements. Operational diagnostics report acceptance, token cost, runtime, verifier calls and repair calls, while embedding based novelty checks examine similarity to the reference corpus. Most importantly, blind expert review evaluates mathematical correctness, curriculum alignment, difficulty match, clarity and final readiness under a teacher rubric. The results show high correctness and alignment, but only about 56\% of completed review records satisfy the full readiness criterion. This gap is a useful result rather than only a weakness: it shows that automated verification can improve reviewability, but professional educational judgment remains necessary before classroom or examination use. The need for expert checks also accords with the relevance and wording limitations reported by \citet{cassano2026instructors}. The evaluation settings and rubrics differ, so these findings are a design comparison rather than a direct performance benchmark.

\subsection{Quality Gaps and Current Boundaries}
\label{sec:discussion-gaps}
The remaining errors clarify what the current system does not yet solve. The most frequent expert marked issue is an invalid problem statement. This can occur when a draft has fluent wording and plausible formulas but still lacks a necessary condition, describes an impossible configuration or fails to guarantee a unique answer. Formula and formatting issues also remain because mathematical notation is sensitive to rendering, symbol placement and local examination conventions. Missing or incorrect solution steps indicate that a model may reach the intended answer while presenting reasoning that is incomplete, mathematically unjustified or unsuitable for the expected high school method.

Several validity boundaries should therefore be recognized. First, the Recognition, Comprehension and Application labels are teacher operational categories used in the studied examination setting. They are not calibrated psychometric item parameters. Second, the verifier uses a separate role and prompt, but it is not a fully independent assessment authority. Verifier acceptance should be treated as evidence of improved reviewability, not as proof of mathematical or pedagogical readiness. Third, the empirical evaluation is limited to Vietnamese Grade 12 mathematics and the three assessment schemas in Table \ref{tab:exam_section_schema}. Adapting VeriMath to another educational system would require replacing or contextualizing the curriculum metadata, schema definitions, anchor corpus, difficulty conventions, annotation protocol and expert rubric. The architecture permits such configuration, but whether comparable performance transfers across grades, subjects or countries remains an empirical question.

The empirical evaluation also has practical limits. It does not include controlled ablation experiments that isolate the individual effects of Difficulty Anchoring, self-improvement, CVM or formatting repair. It does not compare the method against retrieval only, prompt only or teacher authored baselines under the same review protocol. Expert review is a strength because it is blind and rubric based, but the number of completed review records is still limited. The study also does not measure student learning gains, test equating, long term teacher workload reduction or performance in official high stakes examinations. For these reasons, the current system should be used for teacher reviewed draft generation and practice material authoring, not for autonomous official test production.

\subsection{Future Research Directions}
\label{sec:discussion-future}
Future work should strengthen both the research method and the platform function. On the methodological side, controlled ablations can quantify the separate contributions of anchoring, repeated verification, repair prompts and novelty checks. Independent verifier model families, symbolic checks for answer uniqueness, rendering based formula inspection and template leakage detection can make automated review more reliable. On the empirical side, larger expert panels and classroom pilots can evaluate whether the platform reduces teacher workload, improves feedback quality and supports student preparation.

The platform should also evolve from item authoring to a fuller digital assessment environment. Teachers should be able to select topics, assemble tests, deploy them directly to a class, receive automatic grading and inspect intelligent critique for both item quality and student responses. Teacher edits, rejection reasons and student performance logs can then become evidence for future corpus refinement and adaptive practice. This direction supports a closed educational ecosystem in which generation, testing, grading and evaluation inform one another while teachers retain final pedagogical authority.

Another important direction is specialized research for gifted students and advanced mathematical reasoning. Future corpora can select distinct topics, competition styles and examination types for higher difficulty settings, including topic focused tests, school team selection tests and advanced practice examinations. Recent multimodal benchmarks such as MathNet show the value of large, diverse collections for mathematical reasoning and retrieval across topics, languages and visual formats \citep{shaden2026mathnet}. In the Vietnamese setting, this line can guide the development of more challenging assessments and computer based learning environments that go beyond routine national examination preparation.

\section{Conclusion}
This paper presented VeriMath, a configurable human-in-the-loop framework for curriculum-constrained mathematics assessment authoring. The framework separates shared authoring and verification mechanisms from the assessment schemas, curriculum outcomes, difficulty conventions, reference corpus and expert rubric supplied by a deployment context. It combines teacher annotated difficulty anchors with consecutive verification and verifier guided repair, producing drafts for professional review rather than treating model outputs as final assessment items.

We instantiated the framework through a Vietnamese Grade 12 national-curriculum case study using 4,555 teacher annotated reference items and 50 generation requirements. With both evaluated backbones, 47 of 50 requirements reached verifier acceptance. Blind expert review showed high rates of mathematical correctness and curriculum alignment, while 56.8\% and 56.4\% of completed review records satisfied the strict readiness rubric. The main remaining failures involved invalid problem statements, formula or formatting problems and incomplete or incorrect solution steps. These findings clarify both the value and the boundary of automated verification: it can improve the quality and reviewability of generated assessment drafts, but it cannot replace expert teacher judgment.

The broader implication is that responsible artificial intelligence for mathematics assessment should be designed as a teacher-supported authoring environment rather than an autonomous examination generator. The Vietnamese case study supports the feasibility of combining context-specific anchors with repeated verification and repair in one tightly constrained curriculum setting. It does not establish cross-country generalization. Instead, the framework provides a basis for adaptation by making local educational constraints explicit and configurable, while preserving traceability and professional oversight.

Future work should focus on the failure modes identified by expert review. Stronger checks for statement validity, answer uniqueness, mathematical rendering and solution completeness are necessary because many errors are subtle and pedagogically consequential. Controlled ablation studies should measure the individual contributions of anchoring, self improvement, verification and repair. Evaluations with independently constructed corpora, schemas and expert rubrics from other grades, subjects and educational systems are needed to test empirical transfer while keeping educational validity as the central requirement.

\bibliography{references}

\end{document}